\documentclass{ifacconf}

\usepackage{graphicx}      
\usepackage{natbib}        
\usepackage{subfig}
\usepackage{float}
\usepackage{siunitx}
\usepackage[htt]{hyphenat}
\usepackage{algorithm}
\usepackage{algpseudocode}
\usepackage{amsmath}
\usepackage{amsfonts}
\usepackage{amssymb}
\usepackage[acronym]{glossaries}
\usepackage{url} 

\usepackage{afterpage}
\usepackage{stfloats} 

\makeatletter
\newcommand*\dotp{\mathpalette\dotp@{.5}}
\newcommand*\dotp@[2]{\mathbin{\vcenter{\hbox{\scalebox{#2}{$\m@th#1\bullet$}}}}}
\makeatother

\newacronym{ROS}{ROS}{Robot Operating System}
\newacronym{FMM}{FMM}{Fast Marching Method}
\newacronym{FM2}{FM\textsuperscript{2}}{Fast Marching Square}
\newacronym{ETA}{ETA}{Estimated Time of Arrival}
\newacronym{SLAM}{SLAM}{Simultaneous Localization and Mapping}
\newacronym{MPC}{MPC}{Model Predictive Control}
\newacronym{ACO}{ACO}{Ant Colony Optimization}
\newacronym{APF}{APF}{Artificial Potential Field}
\newacronym{LTU}{LTU}{Luleå University of Technology}
\newacronym{UDF}{UDF}{Unsigned Distance Field}

\begin{document}
\begin{frontmatter}

\title{Role-Adaptive Collaborative Formation Planning for Team of Quadruped Robots in Cluttered Environments}

\thanks[footnoteinfo]{This work was partially supported by the Wallenberg AI, Autonomous Systems and Software Program (WASP) funded by the Knut and Alice Wallenberg Foundation.}

\author{Magnus Norén} 
\author{Marios-Nektarios Stamatopoulos} 
\author{Avijit Banerjee}
\author{George Nikolakopoulos}

\address{Robotics and Artificial Intelligence Group, Department of Computer, Electrical and Space Engineering, Lulea University
of Technology, 971 87 Lule\r{a}, Sweden (Corresp. author e-mail: magnus.noren@ltu.se).}

\begin{abstract}                
%
This paper presents a role-adaptive Leader–Follower-based formation planning and control framework for teams of quadruped robots operating in cluttered environments. Unlike conventional methods with fixed leaders or rigid formation roles, the proposed approach integrates dynamic role assignment and partial goal planning, enabling flexible, collision-free navigation in complex scenarios. Formation stability and inter-robot safety are ensured through a virtual spring–damper system coupled with a novel obstacle avoidance layer that adaptively adjusts each agent’s velocity. A dynamic look-ahead reference generator further enhances flexibility, allowing temporary formation deformation to maneuver around obstacles while maintaining goal-directed motion. The \acrfull{FM2} algorithm provides the global path for the leader and local paths for the followers as the planning backbone. The framework is validated through extensive simulations and real-world experiments with teams of quadruped robots. Results demonstrate smooth coordination, adaptive role switching, and robust formation maintenance in complex, unstructured environments. A video featuring the simulation and physical experiments along with their associated visualizations can be found at \url{https://youtu.be/scq37Tua9W4}.

%

\end{abstract}

\vspace{-3mm}

\end{frontmatter}

\section{Introduction}

Formation path planning is an actively researched topic in the field of robotics. It concerns the problem of finding optimal paths for a group of mobile robots through a known or unknown environment, while maintaining a predetermined formation.
Applications include area coverage, disaster relief, traffic monitoring, agricultural sowing, and search and rescue missions (\cite{formation_survey}).
Having a group of robots perform a task collectively increases the overall robustness and provides a safeguard against single points of failure.
In many of these applications, it is crucial to keep the robots within communication range of each other. There are also scenarios in which robots are required to carry a payload together. 
Another specific use case is if each robot is equipped with a different kind of measuring apparatus, and the measurements need to be spatially synchronized.

Concurrently, in recent years, quadruped robots have gained attention for their versatility and superior ability to traverse uneven and obstacle-cluttered terrains compared to traditional wheeled mobile robots (\cite{QuadrupedReview}). Their capability to operate reliably in challenging environments such as rocky, muddy or debris-filled areas makes them suitable for applications in mines, forests, and disaster sites. Additionally, enabling coordinated movement in formation, while maintaining flexibility to adapt or break formation when necessary, can enhance their effectiveness and robustness in dynamic and unstructured environments.
\begin{figure}
    \centering
    \includegraphics[width=0.9\linewidth]{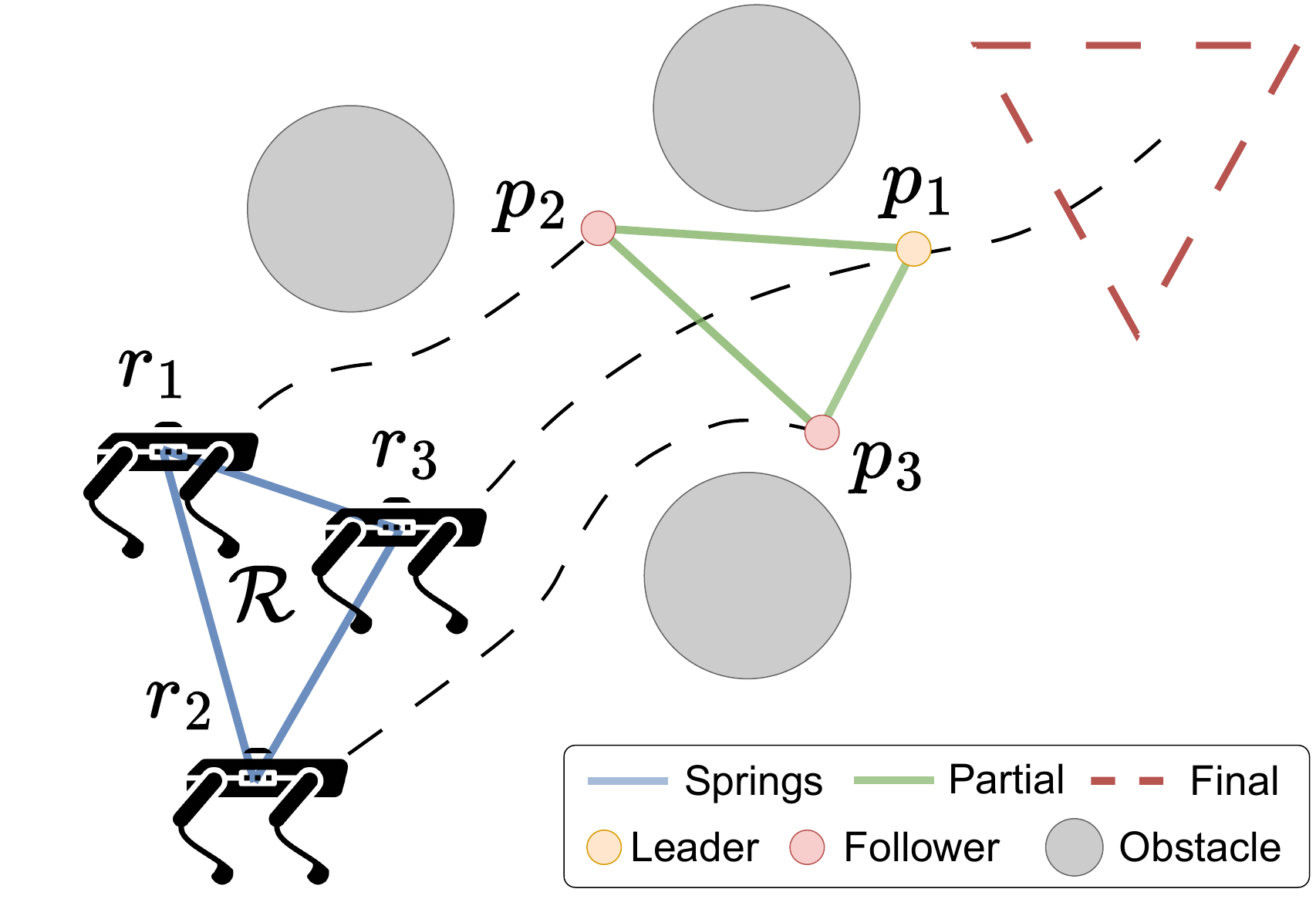}
    \caption{Overview of the proposed framework. The current dynamically assigned leader, $r_3$, plans a global path to the final formation goal. The remaining robots are assigned partial goals $p_2, ... p_N$ and plan local paths to them. The formation is connected by virtual springs and dampers.}
    \label{fig:concept}
    \vspace{-3mm}
\end{figure}
In this paper, we present a framework for formation path planning and control for holonomic ground robots, based on a Leader-Follower architecture augmented with virtual springs and dampers. The proposed planner aims to maintain a user-defined formation shape when moving towards a goal through an obstacle-rich environment. A schematic overview of the concept is shown in Fig. \ref{fig:concept}.

\subsection{Related work}


Formation path planning has been extensively studied in multi-robot systems, often employing a two-layered architecture that separates the tasks of path planning and formation maintenance. Typically, a global path is computed using classical motion planning algorithms, such as grid-based methods or sampling-based planners like RRT or PRM, while a formation control strategy ensures that agents preserve a desired geometric configuration along the path. The most prominent formation control strategies fall into three categories: Virtual Structure, Behavior-Based, and Leader-Follower.

The Virtual Structure approach, exemplified by \cite{VirtualStructures2} and \cite{virtual_structures_zhen2022}, treats the entire formation as a rigid geometric body. A single trajectory is planned for the formation’s reference frame, and individual robot trajectories are derived through fixed geometric transformations relative to this virtual frame. This method ensures high coordination precision and is effective in open environments. However, its rigidity limits flexibility in cluttered or narrow environments, where planning a collision-free path for the entire structure may be infeasible. Additionally, the assumption of constant relative positioning can lead to significant inefficiencies or failures when the environment requires the formation to adapt or deform dynamically.

In contrast, Behavior-Based methods (\cite{behavior2021}, \cite{behavior2024}) take a decentralized, bio-inspired approach in which each robot executes a set of simple, predefined behaviors, such as maintaining inter-agent spacing, avoiding obstacles, and progressing toward a goal. These behaviors are typically combined using arbitration or potential field-based mechanisms. While this method provides scalability and robustness, especially in dynamic and unpredictable environments, it suffers from a lack of centralized coordination. This often results in suboptimal or inconsistent global trajectories, especially in complex environments where locally reactive decisions can lead to dead ends or inefficient paths (\cite{comprehensive_study}).

The Leader-Follower approach (\cite{LeaderFollower1}, \cite{leaderfollower2023},) offers a more structured alternative. In this framework, a path is planned for a designated leader robot, while the remaining agents (followers) employ control laws to maintain a fixed relative position within the formation. Obstacle avoidance is typically handled by allowing followers to temporarily deviate from their nominal positions when necessary, with mechanisms in place to recover the formation once the obstacle is passed. This strategy simplifies coordination and enables formations to traverse more constrained environments. However, it also introduces limitations, for instance, a fixed leader can be inefficient in highly dynamic or maze-like environments where frequent changes in direction may result in poor path quality or collisions.

\cite{FM2_PathPlanning} and \cite{Maritime} combine the Grid Search method \acrfull{FM2} with the Leader-Follower approach. \acrshort{FM2} has the advantage over other classical algorithms of producing safe and smooth trajectories without additional refinement steps. These methods show promising simulation results, but lack validation with physical platforms.

While each approach has its strengths, none fully resolves the trade-off between coordination accuracy, adaptability to dynamic environments, and global path efficiency. This motivates hybrid frameworks and adaptive strategies that can better balance these competing demands.

\afterpage{%
  \clearpage
    \begin{figure*}[b!]
        \centering
        \includegraphics[width=0.75\linewidth]{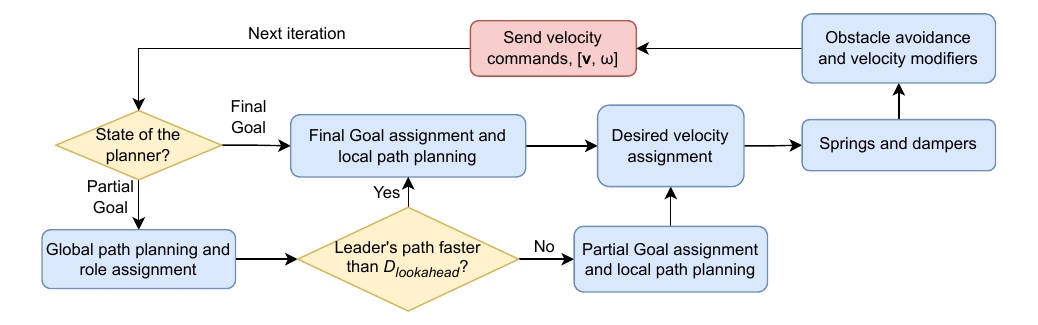}
        \caption{High-level flowchart of the main loop of the proposed formation path planner.}
        \label{fig:flowchart}
    \end{figure*}
}
\subsection{Contributions}


In this paper, a Leader-Follower-based role-adaptive collaborative formation planning framework using \acrshort{FM2} is proposed. The main contributions are as follows.

Firstly, the robots are treated as non-unique in the sense that none of them occupy a predetermined position in the formation, nor is there a fixed leader.
Instead, a dynamic, reactive role adaptation scheme is proposed, enabling the formation to adjust to cluttered environments and complex obstacle geometries by assigning agents to formation positions online. This approach enhances the framework’s responsiveness during motion, allowing seamless interchange of the leading robot along with matching the rest to the formation positions and preventing issues associated with static or outdated role assignments.

Secondly, unlike previous state of the art approaches where partial goals were directly constructed based on the leader’s instantaneous position, potentially resulting in jerky motions and reduced flexibility, the proposed method employs a dynamic “look-ahead” reference generator. This mechanism provides the formation robots with greater adaptability, enabling them to temporarily relax the strict formation shape to navigate around obstacles while progressing toward the final goal.

Thirdly, the proposed planner uses a flexible system of virtual springs and dampers to avoid inter-robot collisions. To prevent the spring forces from pushing the robots into obstacles, we introduce a novel obstacle avoidance layer, which modifies the preliminary command velocity vector based on the value and gradient of an \acrfull{UDF} of the occupancy map.


The framework is validated with formations of three and four Unitree Go1 quadruped robots, both through simulation in Gazebo and extensive evaluation in real-life experiments in different lab-built environments.

\section{Hybrid Leader-Follower formation planning method}

\subsection{Overview}\label{sec:Overview}

Let $\mathcal{R} = \{r_1,...,r_N\}$ be the set of robots. The objective of the formation planner is to find collision-free trajectories for each $r_k$ to a common goal, $\xi_\mathrm{goal}$, while adhering to a prescribed formation shape, $\mathcal{C}$, as closely as possible given environmental constraints. The desired geometry of the formation is described using a base configuration, which is a set of points, $\mathcal{C} = \{\mathbf{c}_1,...,\mathbf{c}_N\}$, where $\mathbf{c}_i \in \mathbb{R}^2$. $\mathbf{c}_1$ is the leader position. The convention of placing the leader along the positive x-axis, which aligns with the predefined forward direction of the formation as a whole, is adopted.
Note that the positions $\mathbf{c}_i$ are not associated with any particular robots, but rather describe a "role" in the formation. The dynamic role assignment feature assigns and reassigns roles to the robots to adapt to the current circumstances. This can be formally described as a dynamically updated bijection from $\mathcal{R}$ to $\mathcal{C}$.


In the proposed planner a path for each robot is calculated based on the \acrfull{FM2} (\cite{FM2_PathPlanning}) path planner,
an extension of the \acrfull{FMM} (\cite{FastMarching}). The main benefits of using \acrshort{FMM} or \acrshort{FM2} in path planning are completeness (i.e., if there exists a path, the algorithm is guaranteed to find it), absence of local minima, optimal trajectories and fast computation (\cite{4D_FM2}). 

\begin{figure}[H]
    \centering
    \includegraphics[width=\linewidth]{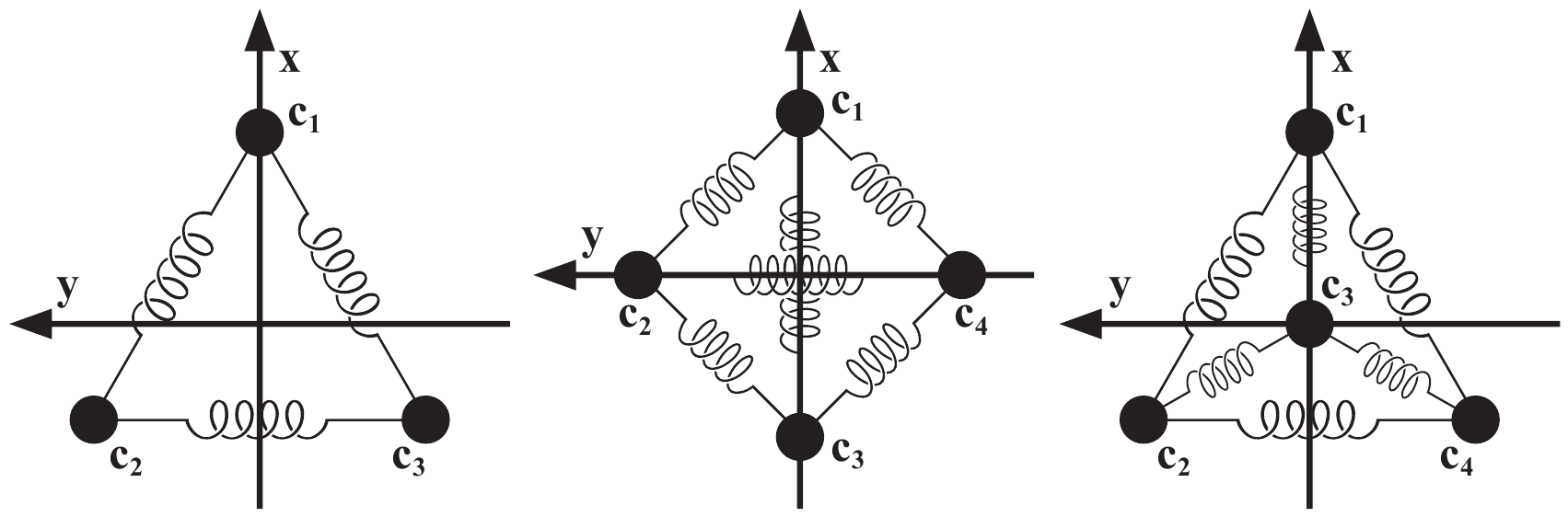}
    \caption{Examples of base configurations and connection setups for formations with three or four robots.}
    \label{fig:formation_examples}
\end{figure}

In addition, to keep the robots within close range, and simultaneously avoid inter-robot collisions, virtual springs and dampers are defined as a set of connections, $\mathcal{S} = \{s_1,...,s_M\}$, where each one connects any two points in $\mathcal{C}$ (i.e., they are associated with a pair of roles rather than a specific pair of robots). Each connection has its own properties, detailed in Section \ref{sec:springs_dampers}, allowing them to differ across connections. Fig. \ref{fig:formation_examples} shows a few examples of base configurations and connection setups for formations with three or four robots.


When launching the framework, the planner starts in an inactive state. As a formation goal, $\xi_\mathrm{goal} \in \mathrm{SE}(2)$, is fed to the planner in the form of a pose for the formation center, the planner enters an active state (if it was already active, the new formation goal replaces the old one). Individual robot goals, $\mathbf{x}_{\mathrm{goal},i}$, are calculated by transforming the base configuration to this pose. While active, the planner can be in either the Partial Goal state, in which a global path is planned from the leader to $\xi_\mathrm{goal}$, and local paths are planned for the followers to intermediate goals along the way, or the Final Goal state, in which there is no leader, and local paths are planned from each robot to their assigned $\mathbf{x}_{\mathrm{goal},i}$. Upon receiving a new formation goal, the planner starts out in the Partial Goal state. A high-level flowchart of the main loop is shown in Fig. \ref{fig:flowchart}.


The formation path planner includes several tuning parameters, denoted with a tilde (e.g., $\tilde{v}_{\mathrm{max,x}}$) to distinguish them from other variables.

\subsection{Fast Marching Square}\label{sec:FM2}

\subsubsection{The Fast Marching Method}

Conceptually, the \acrlong{FMM} calculates the time of arrival function, $D(\mathbf{x}), \mathbf{x}\in\mathbb{R}^n$, of a wavefront propagating from a set of starting points in $n$ dimensions, given a propagation velocity function, $W(\mathbf{x})$. In practice, both $D(\mathbf{x})$ and $W(\mathbf{x})$ are represented as discrete n-dimensional grids. Since we are dealing with a two-dimensional application, the method will only be described in two dimensions. The notation $D(i,j)$ and $W(i,j)$, where $i$ and $j$ are cell indices in the $x$ and $y$ directions respectively, will be used when referring to the discrete grids.

The propagation of wavefronts is governed by the Eikonal equation,

\begin{equation}
    |\nabla D(\mathbf{x})|W(\mathbf{x}) = 1.
    \label{ch2:eq:eikonal}
\end{equation}

The \acrshort{FMM} finds an approximate solution to \eqref{ch2:eq:eikonal} on the discrete grids. In case $W(\mathbf{x})$ is constant (i.e., representing a homogeneous medium), the solution, $D(\mathbf{x})$ will represent the closest euclidean distance to any point in the starting set, in other words, an \acrshort{UDF} of $W(\mathbf{x})$.
%
%
%

\subsubsection{Fast Marching used in path planning}\label{sec:FMM_planning}
%
%

To use the \acrshort{FMM} as a path planning algorithm, we start with a binary occupancy grid describing the environment, inflated to account for the maximum radius, $\tilde{r}_\mathrm{max}$, of the robot. The inflated grid is used as the velocity map, $W(i,j)$. A wavefront is then propagated from the goal point, resulting in a time of arrival map, $D(i,j)$. The path is calculated by performing gradient descent on $D(i,j)$ from the starting point.

The resulting path is optimal in terms of total length. However, it is not very practical for most robotics applications because it contains sharp turns and comes unnecessarily close to obstacles. The \acrlong{FM2} method is designed to remedy these shortcomings. It uses the \acrshort{FMM} twice, hence the "Square" designation. \acrshort{FM2} is performed in the following steps.

\begin{figure*}
    \centering
    \includegraphics[width=0.7\linewidth]{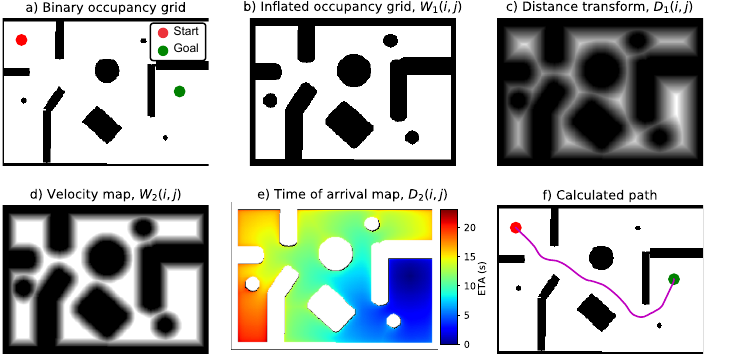}
    \caption{Path planning with the \acrshort{FM2} method. The binary occupancy grid (a) is inflated to occupancy grid $W_1$ (b), from which distance transform $D_1$ is calculated using \acrshort{FMM} (c). $D_1$ is saturated at a safe distance, giving the velocity map $W_2$ (d). Applying \acrshort{FMM} to $W_2$, produces the time of arrival map $D_2$ (e), from which the path is calculated using gradient descent (f).}
    \label{ch2:fig:FM2_planning}
\end{figure*}

We start from an inflated occupancy grid, $W_1(i,j)$ (Fig. \ref{ch2:fig:FM2_planning}a-b). A Fast Marching wavefront is propagated using all the occupied cells as starting set, thus performing a distance transform of $W_1(i,j)$. The resulting time of arrival map, $D_1(i,j)$ (Fig. \ref{ch2:fig:FM2_planning}c), now representing the shortest distance to any obstacle, is saturated at a safe distance, $d_\mathrm{safe}$, chosen as a distance considered safe for the robot to maneuver at full speed, and then normalized to the interval [0, 1] (Fig. \ref{ch2:fig:FM2_planning}d). The result is used as the velocity map, $W_2(i,j)$, in a second Fast Marching step. In this second step, the wavefront is propagated from the goal point. The path is calculated using gradient descent on the second time of arrival map, $D_2(i,j)$, just as in the standard Fast Marching approach (Fig. \ref{ch2:fig:FM2_planning}e-f).

A path generated by FM2 converges to the time-optimal solution as the cell size approaches zero, assuming the robot follows a velocity profile proportional to the velocity map. In the following, the values of $D_2(i,j)$ are referred to as the Estimated Time of Arrival (ETA), as the continuously updated paths may not correspond exactly to the actual arrival times.
It is given in units of normalized time, i.e., the time in seconds it would take a robot travelling at a speed proportional to the velocity map, and a maximum speed of \SI{1}{m/s}, to reach the goal.

\subsubsection{Velocity maps}

Since the leader's path determines the behaviour of the formation as a whole, it requires more clearance than those of the followers. For this reason, two variants of the velocity map are generated, with the only difference being the saturation distance (or safe distance). The velocity map for planning the leader's path will be referred to as $W_\mathrm{A}(i,j)$, and the one for followers as $W_\mathrm{B}(i,j)$. The latter one is used for all robots when they are sufficiently close to the formation goal (i.e., in the Final Goal state), as the concept of a leader no longer applies. The two safe distances are given by $\tilde{d}_\mathrm{safe,A}$ and $\tilde{d}_\mathrm{safe,B}$ respectively.

\subsection{Global path planning and role assignment}\label{sec:role_assignment}

In the Final Goal state, this step is bypassed (see Fig. \ref{fig:flowchart}), and the planner goes directly to final goal assignment, detailed in section \ref{sec:goal_assignment}.
Otherwise, an \acrshort{FM2} path is calculated from each robot's position to the formation goal center on $W_\mathrm{A}(i,j)$ (the velocity map with the larger saturation distance). The robot with the fastest path in terms of \acrshort{ETA} is normally assigned the leader role.
However, to avoid situations where the leadership switches back and forth between two competing robots, a hysteresis is added, i.e., a candidate robot's path must be faster than the current leader's by $\tilde{D}_\mathrm{switch}$ to reassign leadership.



The follower robots are assigned formation roles so that the angular order with respect to the leader robot from the base configuration if maintained. For larger formations where the angular order is ambiguous, they are first grouped by their current distance to the leader.

The leader path's \acrshort{ETA} is compared to $\tilde{D}_\mathrm{lookahead}$. If it is shorter, the planner enters the Final Goal state, where it remains until all robots have reached their individual goals.


\subsection{Goal assignment and local path planning}\label{sec:goal_assignment}

The goal assignment is handled differently depending on whether the planner is in the Partial Goal state or the Final Goal state.

\subsubsection{Partial Goal state}
The leader's partial goal, $\mathbf{p}_1$, is defined as the point located $\tilde{D}_\mathrm{lookahead}$ units of normalized time (see section \ref{sec:FM2}) along its path. 
The partial goals for the followers, $\mathbf{p}_2,...,\mathbf{p}_N$ are calculated as

\vspace{-3mm}
\begin{equation}
    \mathbf{p}_i = \mathbf{R}(\theta_{\mathbf{p}_1})(\mathbf{c}_i - \mathbf{c}_1) + \mathbf{p}_1,
\end{equation}

where $\mathbf{R}(\theta_{\mathbf{p}_1})$ is the rotation matrix corresponding to the leader path direction at $\mathbf{p}_1$ (see Fig. \ref{fig:concept}). It may happen that the follower partial goals constructed this way end up inside of, or close to, an obstacle. In this case, they are iteratively shifted towards $\mathbf{p}_1$ in small steps, until the grid cell they reside in satisfies $W_\mathrm{B}(i,j)\geq \tilde{W}_\mathrm{min,partial}$, or until they coincide with $\mathbf{p}_1$, whichever comes first.
%



\subsubsection{Final Goal state}
When the leader robot has a path to the formation goal center which is shorter than $\tilde{D}_\mathrm{lookahead}$, the planner enters the Final Goal state. The final goal assignment is the bijection $f: \mathcal{R} \leftrightarrow \mathcal{C}$ that minimizes the sum of squared distances between robots and their final goals. The assignment is reevaluated at each iteration after it is triggered, and the roles of the robots (i.e., their positions in the base configuration) are updated accordingly. 

\acrshort{FM2} paths are planned for all robots from their positions to their assigned (partial or final) goals on $W_\mathrm{B}(i,j)$, except for the leader (if applicable), which retains its previously computed path.
%
Each robot is assigned a "desired velocity" vector, $\mathbf{v}_{\mathrm{des},k}$, for $k\in[1,N]$. The direction of $\mathbf{v}_{\mathrm{des},k}$ is the initial direction of the robot's path.
The magnitude of $\mathbf{v}_{\mathrm{des},k}$ is set to the maximum allowed velocity for the robot.

\subsection{Springs and dampers}\label{sec:springs_dampers}


To maintain proximity and prevent inter-robot collisions, virtual springs and dampers are defined as connections $\mathcal{S} = {s_1,\dots,s_M}$, each linking a pair of points in $\mathcal{C}$ and associated with roles rather than specific robots. For each $s_i \in \mathcal{S}$, the role assignment specifies the connected robots. These virtual elements are inspired by, but not equivalent to, physical springs and dampers; they modify the desired velocity output directly rather than applying forces.

The spring "force" for $s_i$ is calculated as
\begin{equation}
    \small
    \Delta v_{\mathrm{spring},i} =
    \begin{cases}
         -\tilde{k}_{\mathrm{rep},i}(l_i-\tilde{l}_{0,i})^2, &\textrm{if}\ l_i<\tilde{l}_{0,i}, \\
        \min(\tilde{k}_{\mathrm{att},i}(l_i-\tilde{l}_{0,i})^2, \Delta \tilde{v}_{\mathrm{max,att},i}), &\textrm{if}\ l_i \geq \tilde{l}_{0,i},    
    \end{cases}
\end{equation}

where $\tilde{l}_{0,i}$ is the relaxed length of $s_i$, $l_i$ is the current length of $s_i$ (i.e., the distance between the robots connected by $s_i$), $\tilde{k}_{\mathrm{rep},i}$ is the repulsive spring stiffness, $\tilde{k}_{\mathrm{att},i}$ is the attractive spring stiffness (typically chosen lower than $\tilde{k}_{\mathrm{rep},i}$ to allow the formation to deform when needed) and $\Delta \tilde{v}_{\mathrm{max,att},i}$ is the maximum magnitude by which the spring can modify the velocity when extended (preventing the attractive force from growing without bounds when the connected robots are far from each other).

The damping "force" for $s_i$ is calculated as
\begin{equation}
    \Delta v_{\mathrm{damp},i} =
    \begin{cases}
        \tilde{b}_{\mathrm{att},i}\frac{\Delta l_i}{\Delta t},\qquad &\textrm{if}\ \Delta l_i < 0, \\
        \tilde{b}_{\mathrm{rep},i}\frac{\Delta l_i}{\Delta t},\qquad &\textrm{if}\ \Delta l_i \geq 0, 
    \end{cases}
\end{equation}

where $\Delta l_i=l_i-l_{\mathrm{prev},i}$, $\Delta t=1 / \tilde{f}$ is the time since the previous iteration, $\tilde{b}_{\mathrm{att},i}$ is a damping coefficient for when the connected robots are approaching each other and $\tilde{b}_{\mathrm{rep},i}$ is a damping coefficient for when they are drifting apart. In case $s_i$ connects different robots than in the previous iteration (because the roles have just been reassigned), the damping step is omitted.

After all spring and damping forces have been calculated, the (preliminary) velocity vector $\mathbf{v}_k$ is introduced for each $r_k, k\in[1,N]$. Let $\mathcal{S}_k$ be the subset of $\mathcal{S}$ that connects to $r_k$. Then
\begin{equation}
    \mathbf{v}_k = \mathbf{v}_{\mathrm{des},k} + \sum_{\{i:s_i\in\mathcal{S}_k\}} \left( \Delta v_{\mathrm{spring},i} + \Delta v_{\mathrm{damp},i}\right) \hat{\mathbf{v}}_{ki},
\end{equation}

where $\hat{\mathbf{v}}_{ki}$ is a unit vector pointing from $\mathbf{x}_k$ to the other robot connected through connection $s_i$.

\subsection{Obstacle avoidance}


After applying the springs and dampers, an additional obstacle avoidance step is necessary to ensure that the robots are not driven into obstacles. 
At a high level, this step ensures that a robot already near an obstacle is restricted from moving significantly closer. It is carefully designed to not completely negate the effect of the springs, since in particular the repulsive spring force serves an important purpose in preventing inter-robot collisions.

If the value of $W_\mathrm{B}(i,j)$ at the position, $\mathbf{x}_k$, of $r_k$ satisfies $W_\mathrm{B}(i,j) < 1$ and $\nabla W_\mathrm{B}(\mathbf{x}_k) \dotp \mathbf{v}_k < 0$ (indicating that the preliminary velocity vector moves the robot closer to the obstacle), the obstacle avoidance is triggered.
The gradient, $\nabla W_\mathrm{B}(\mathbf{x}_k)$, is approximated by applying the Sobel operator to a 3x3 patch of $W(i,j)$, centered around the point of interest.

\begin{figure}[h]
    \centering
    \includegraphics[width=0.8\linewidth]{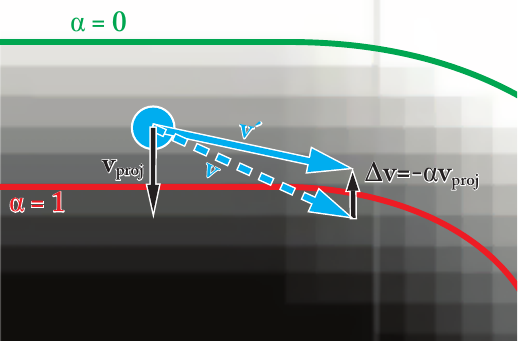}
    \caption{Illustration of the obstacle avoidance step. The blue circle is the robot's position. $\mathbf{v}$ denotes the velocity vector before the adjustment, and $\mathbf{v'}$ denotes the velocity vector after the adjustment. The adjustment factor in this case is approximately $\alpha=0.6$.}
    \label{fig:obstacle_avoidance}
\end{figure}




By construction, the gradient of the velocity map typically satisfies $|\nabla W_\mathrm{B}(i,j)| \approx \tilde{h} / \tilde{d}_\mathrm{safe,B}$ at points where $W_\mathrm{B}(i,j) < 1$, where $\tilde{h}$ is the cell size. If the magnitude of the gradient is significantly lower (half of this value is used as a threshold in the implementation), it indicates that the robot is at a ridge or saddle point of $W_\mathrm{B}(i,j)$, in which case the step is canceled. Otherwise, let $\mathbf{v}_{\mathrm{proj},k}$ be the component of $\mathbf{v}_k$ parallel to $-\nabla W_\mathrm{B}(i,j)$. An updated preliminary velocity, $\mathbf{v}'_k$, is calculated according to \eqref{eq:obstacle_avoidance},

\vspace{-2mm}
\begin{equation}
    \mathbf{v}'_k = \mathbf{v}_k - \alpha \mathbf{v}_{\mathrm{proj},k},
    \label{eq:obstacle_avoidance}
\end{equation}

where $\alpha$ is an adjustment factor found by linearly remapping the value of $W_\mathrm{B}(i,j)$ from the interval [$\tilde{W}_\mathrm{min,avoid}$, 1.0] to [1.0, 0.0] (at values below $\tilde{W}_\mathrm{min,avoid}$, $\alpha$ remains at 1.0).
Fig. \ref{fig:obstacle_avoidance} shows a visual explanation of the obstacle avoidance step. Algorithm \ref{alg:avoid}, in which $i$ and $j$ denote the grid indices of $\mathbf{x}_k$, is the pseudo-code for the step.

\begin{algorithm}
\caption{The obstacle avoidance step}\label{alg:avoid}
\begin{algorithmic}\small
    \State $\mathrm{P^B_{i,j}} \gets \mathrm{get3x3Patch(W_B,i,j)}$
    \State $\nabla W_\mathrm{B}[i,j] \gets \mathrm{sobelGradient(P^B_{i,j})}$
    \If{($W_\mathrm{B}[i,j] < 1)\ \mathbf{and}\ (\nabla W_\mathrm{B}[i,j] \dotp \mathbf{v}_k < 0)\ \mathbf{and}\ (|\nabla W_\mathrm{B}[i,j]|>0.5\ \tilde{h}/ \tilde{d}_\mathrm{safe,B})$}
        \State $\mathbf{v}_{\mathrm{proj},k} \gets \dfrac{\nabla W_\mathrm{B}[i,j] \dotp \mathbf{v}_k}{|\nabla W_\mathrm{B}[i,j]|^2}\nabla W_\mathrm{B}[i,j]$
        \State $\alpha \gets \mathrm{linearMap}(W_\mathrm{B}[i,j], [\tilde{W}_\mathrm{min,avoid}, 1.0], [1.0, 0.0])$
        \State $\mathbf{v}'_k \gets \mathbf{v}_k - \alpha\ \mathbf{v}_{\mathrm{proj},k}$
    \Else
        \State $\mathbf{v}'_k \gets \mathbf{v}_k$
    \EndIf
\end{algorithmic}
\end{algorithm}

\subsection{Velocity modifiers}


The final velocity vector, $\mathbf{v}^*_k$, is given by \eqref{eq:final_vel},
\vspace{-2mm}
\begin{equation}
    \mathbf{v}^*_k = \min\left(|\mathbf{v}'_k|, v_{\mathrm{max,\theta},k}, v_{\mathrm{max,W},k}, v_{\mathrm{max,goal},k}\right)\frac{\mathbf{v}'_k}{|\mathbf{v}'_k|}.
    \label{eq:final_vel}
\end{equation}

\begin{figure*}
    \centering
    \includegraphics[width=0.8\linewidth]{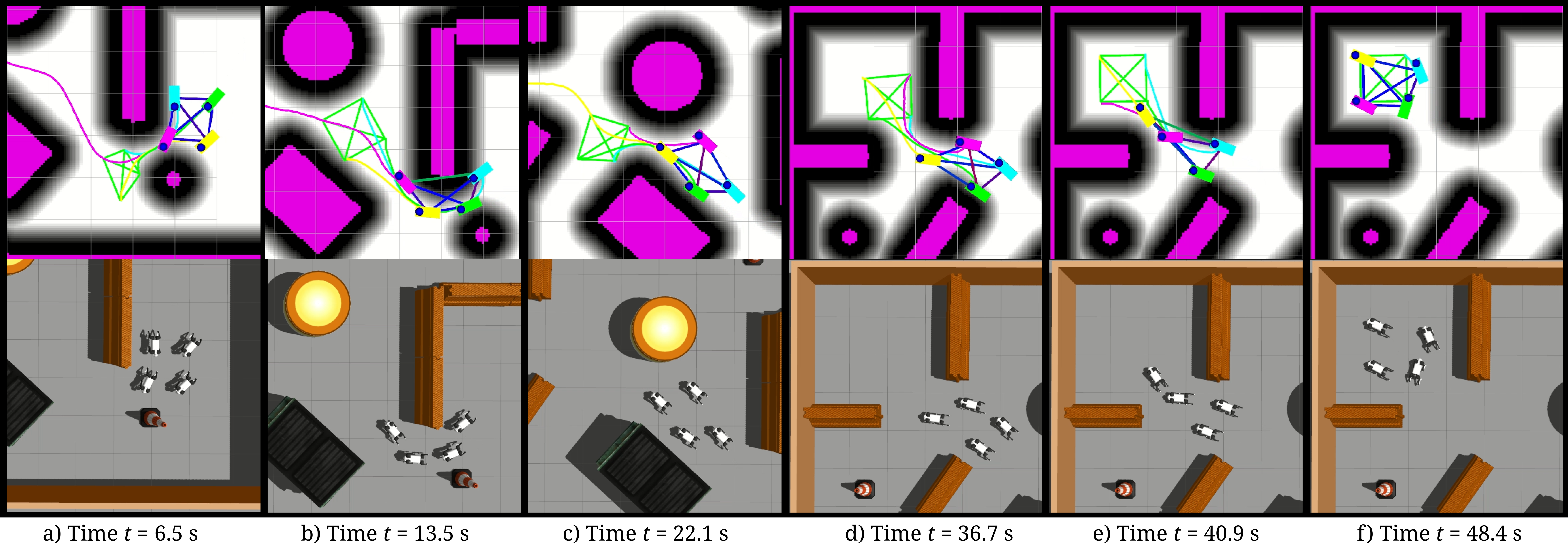}
        \vspace{-3mm}
        \caption{Snapshots from the Gazebo simulation along with RViz visualization.}
        \label{fig:sim_seq}
\end{figure*}

The quantity $v_{\mathrm{max,\theta},k}$ has the purpose of limiting the speed based on the direction the robot is moving. Often, holonomic robots such as the quadrupeds used for validating the planner are able to maintain stability at higher longitudinal than lateral velocities and accelerations. The maximum allowed speed in the longitudinal and lateral directions for $r_k$ are given by the parameters $\tilde{v}_{\mathrm{max,x},k}$ and $\tilde{v}_{\mathrm{max,y},k}$ respectively. Let $\theta_k$ be the body frame angle of $\mathbf{v}'_k$. Then, $v_{\mathrm{max,\theta},k}$ is the radius at angle $\theta_k$ of an ellipse with $\tilde{v}_{\mathrm{max,x},k}$ and $\tilde{v}_{\mathrm{max,y},k}$ as semi-major and semi-minor axes. 
This is calculated using \eqref{eq:direction_speed},
\begin{equation}
    v_{\mathrm{max,\theta},k}=\frac{\tilde{v}_{\mathrm{max,x},k}\ \tilde{v}_{\mathrm{max,y},k}}{\sqrt{(\sin\theta_k\ \tilde{v}_{\mathrm{max,x},k})^2 + (\cos\theta_k\ \tilde{v}_{\mathrm{max,y},k})^2}}.
    \label{eq:direction_speed}
\end{equation}


$v_{\mathrm{max,W},k}$ is found by linearly remapping the value of $W_\mathrm{B}(i,j)$ at the robot's position from the interval [0.0, 1.0] to [$\tilde{v}_{\mathrm{min},k}$, $\tilde{v}_{\mathrm{max,x},k}$]. $\tilde{v}_{\mathrm{min},k}$ is a parameter with the purpose of not stopping completely when close to obstacles, since the high-level controller in the robot might not accept too small velocity commands.

If the planner is in the Final Goal state, $v_{\mathrm{max,goal},k}$ is calculated as
\vspace{-2mm}
\begin{equation}
    v_{\mathrm{max,goal},k} = \frac{||\mathbf{x}_k - \mathbf{x}_{\mathrm{goal},k}||}{\tilde{d}_{\mathrm{slowdown},k}}\ \tilde{v}_{\mathrm{max,x},k}.
\end{equation}

The yaw rate, $\omega_k$, is calculated to attempt to align the robot with its velocity vector. 
%
%
Finally, velocity command consisting of $\mathbf{v}^*_k$ and $\omega_k$ is sent to each robot. The algorithm then goes idle until the next iteration.

\section{Results}

The proposed framework was evaluated in a simulated environment and then deployed in different scenarios in real-life experiments. The evaluation platform is the Go1 legged robot from Unitree Robotics, measuring 0.65x0.28x0.40 m. with a stated maximum speed of \SI{3.5}{m/s}.
The tuning parameters which were kept identical between simulations and lab experiments were given the following values: $\tilde{h}=\SI{0.05}{m}$, $\tilde{d}_\mathrm{safe,A}=\SI{1.50}{m}$, $\tilde{d}_\mathrm{safe,B}=\SI{0.50}{m}$, $\tilde{D}_\mathrm{switch}=\SI{0.05}{s}$, $\tilde{d}_\mathrm{slowdown}=\SI{0.40}{m}$, $\tilde{W}_\mathrm{min,partial}=0.70$, $\tilde{W}_\mathrm{min,avoid}=0.50$, $\tilde{v}_\mathrm{max,x}=\SI{0.50}{m/s}$, $\tilde{v}_\mathrm{max,y}=\SI{0.20}{m/s}$, $\tilde{v}_\mathrm{min}=\SI{0.50}{m/s}$. The planner was running at a rate of \SI{20}{Hz}.
To fully take part of the results, we encourage the reader to watch the video containing both the simulations and experimental results at \url{https://youtu.be/scq37Tua9W4}.
 
\subsection{Simulation results}

A map of size 14x10 m containing a varied assortment of obstacles was constructed in the Gazebo simulation environment. The simulation uses the CHAMP open source development framework for quadruped robots (\cite{Champ}), which contains description files for the Unitree Go1 model. The formation was defined as four robots forming a square of side length 1.0 m like the one shown in the center of Fig. \ref{fig:formation_examples}.
The look-ahead value chosen for the simulation was $\tilde{D}_\mathrm{lookahead}=\SI{5.0}{s}$. The four sides of the formation square had the following connection properties: $\tilde{l}_0=\SI{1.0}{m}$, $\tilde{k}_\mathrm{rep}=\SI{4.5}{m^{-1}s^{-1}}$, $\tilde{k}_\mathrm{att}=\SI{1.25}{m^{-1}s^{-1}}$, $\Delta\tilde{v}_\mathrm{max,attr}=\SI{0.4}{m/s}$, $\tilde{b}_\mathrm{att}=0.1$, $\tilde{b}_\mathrm{rep}=0.04$. The diagonals of the square had the following connection properties: $\tilde{l}_0=\SI{1.41}{m}$, $\tilde{k}_\mathrm{rep}=\SI{3.0}{m^{-1}s^{-1}}$, $\tilde{k}_\mathrm{att}=\SI{0.9}{m^{-1}s^{-1}}$, $\Delta\tilde{v}_\mathrm{max,attr}=\SI{0.3}{m/s}$, $\tilde{b}_\mathrm{att}=0.1$, $\tilde{b}_\mathrm{rep}=0.04$.
\begin{figure}[H]
    \centering
    \includegraphics[width=\linewidth]{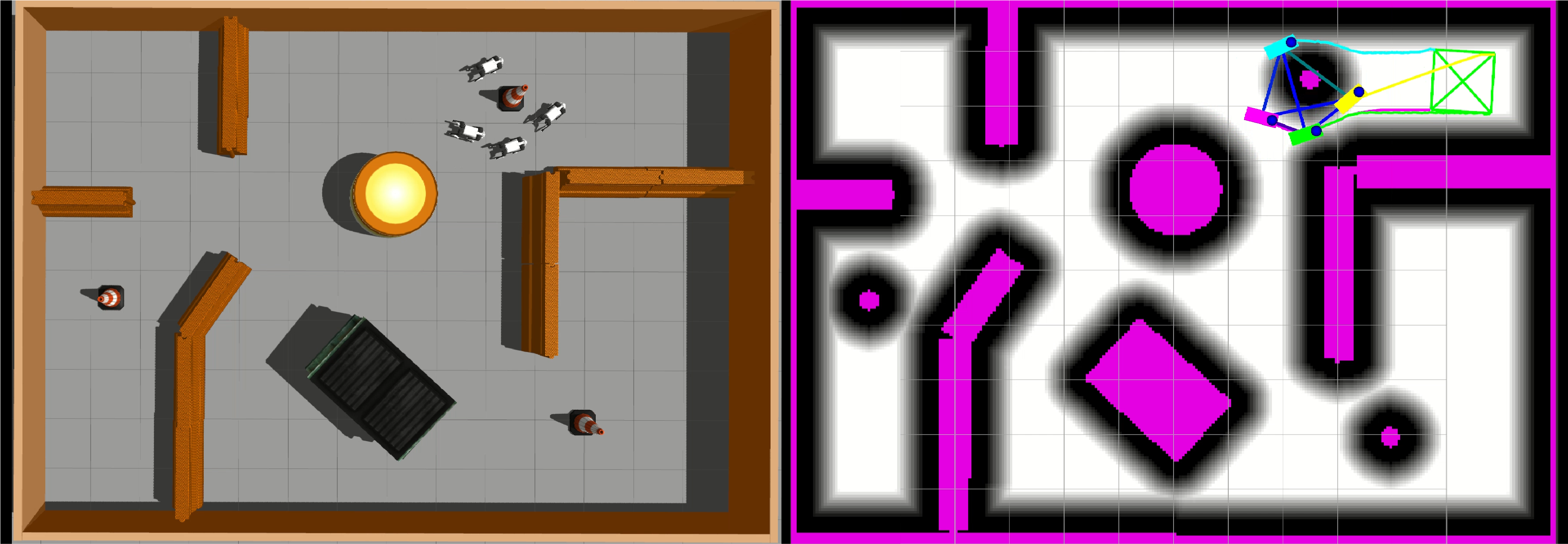}
    \caption{Simulation snapshot highlighting the ability to take different paths around smaller obstacles.}
    \label{fig:sim_split_path}
\end{figure}

\begin{figure*}
    \centering
    \includegraphics[width=\linewidth]{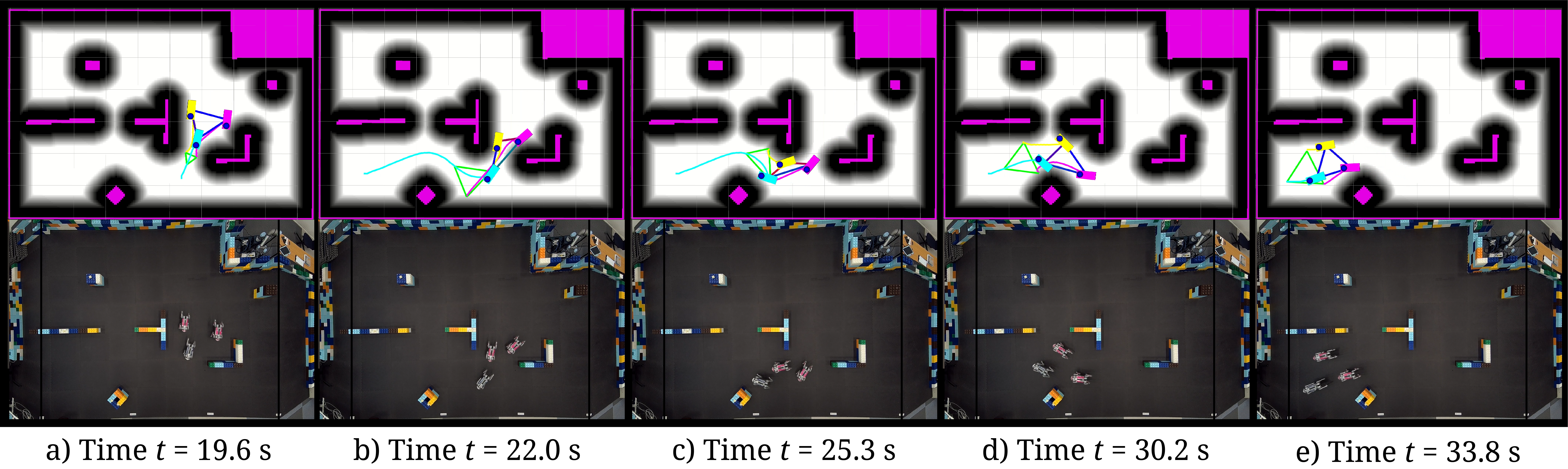}
    \vspace{-5mm}    
    \caption{Snapshots from a lab experiment with an unstructured environment.}
    \label{fig:lab_exp_1}
\end{figure*}

\begin{figure*}[b]
    \centering
    \includegraphics[width=0.88\linewidth]{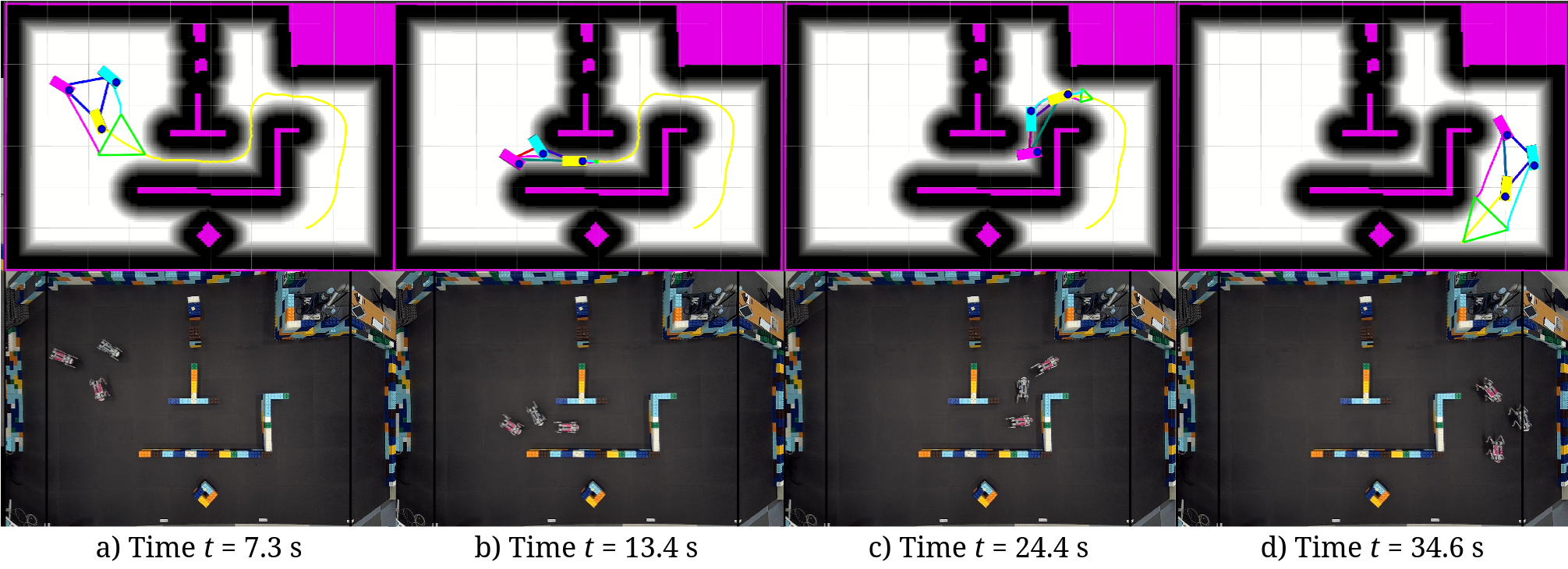}
    \vspace{-3mm}
    \caption{Snapshots from a lab experiment in which the formation is passing through a narrow corridor.}
    \label{fig:corridor}
\end{figure*}

Fig. \ref{fig:sim_seq} shows a sequence of snapshots from the simulation. The sequence demonstrates the deformability of the formation in the presence of obstacles. In particular, in subfig. b) and e) the available space is insufficient for traversing in the prescribed square formation, and the framework reacts by adopting a more flexible shape. The dynamic role assignment is also evident. Up until subfig. b), the robot visualized in purple is the leader, as can be seen by the color of the global path, while from subfig. c) onward, the leadership has been reassigned to the yellow robot. Furthermore, the relative positions of the yellow and purple robots are exchanged after passing through the gap in subfig. d) through f). In subfig. e), the planner has entered the Final Goal state, and in subfig. f), the robots' positions have almost converged to their final goals.
In Fig. \ref{fig:sim_split_path}, the formation is moving past a small obstacle represented by a cone. In this situation, it is advantageous for the robot visualized in cyan to take a different path past the obstacle than the leader, which is made possible through the look-ahead feature of the planner. 
As a result, the formation successfully navigates around the obstacle, demonstrating the adaptability and reactiveness of the proposed framework.

\subsection{Experimental results}
\vspace{-2mm}
The proposed framework was evaluated in a laboratory equipped with a Vicon motion capture system, which provided precise localization for the robots. A centralized planner running on a lab computer generated velocity commands that were transmitted to the robots via WiFi. The experimental area measured 9.5 × 6.5 m, and the formation consisted of three robots arranged in an equilateral triangle with a side length of 1.15 m.
The look-ahead value used in the lab experiments was $\tilde{D}_\mathrm{lookahead}=\SI{2.5}{s}$. The connections had the following properties: $\tilde{l}_0=\SI{1.15}{m}$, $\tilde{k}_\mathrm{rep}=\SI{4.0}{m^{-1}s^{-1}}$, $\tilde{k}_\mathrm{att}=\SI{2.0}{m^{-1}s^{-1}}$, $\Delta\tilde{v}_\mathrm{max,attr}=\SI{0.50}{m/s}$, $\tilde{b}_\mathrm{att}=0.10$, $\tilde{b}_\mathrm{rep}=0.10$.

Various unstructured environments were constructed from basic building blocks and extensive testing of the framework in such environments was performed. Over the course of these experiments, the formation reliably navigated between its waypoints without collisions or other failures. A sequence of snapshots from one of the experiments with an unstructured environment is shown in Fig. \ref{fig:lab_exp_1}.
Initially, the formation maintains its base configuration (subfigure a). As the robots approach and navigate through the obstacles (subfigures b and c), the formation shape adapts to the surroundings, allowing safe passage through the constrained environment. Finally, upon reaching the goal region, the formation restores its original configuration (subfigure e), demonstrating its ability to adapt and re-form as required by the environment.

\begin{figure}[h]
    \centering
    \includegraphics[width=0.85\linewidth]{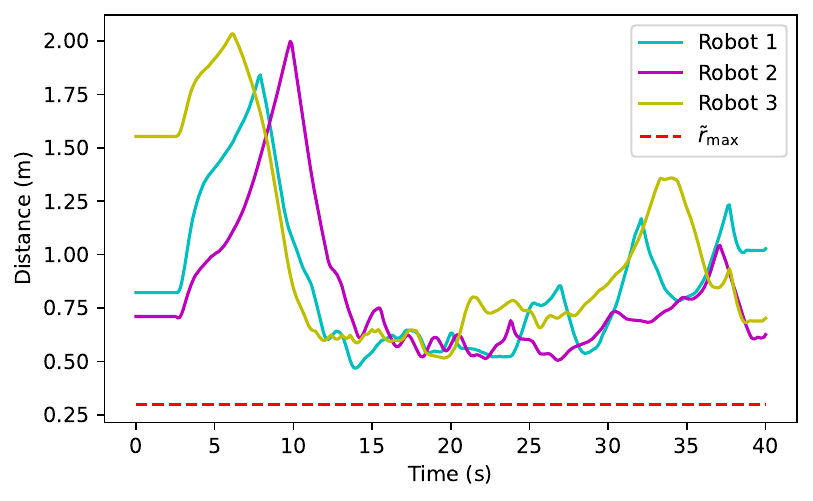}
    \vspace{-3mm}
    \caption{Distances between each robot and its closest obstacle for the narrow corridor lab experiment.}
    \label{fig:robot_obstacle_dist}
\end{figure}

\begin{figure}[h]
    \centering
    \includegraphics[width=0.85\linewidth]{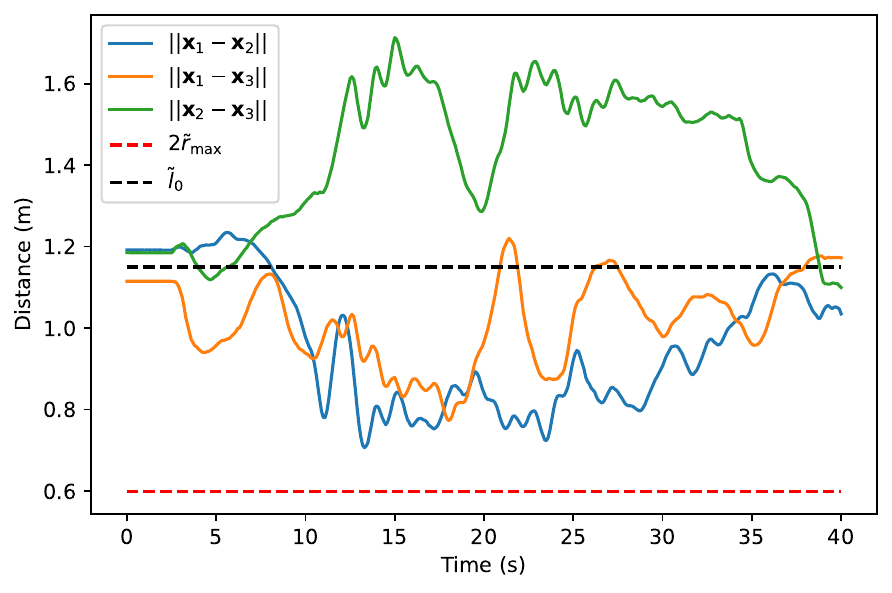}
    \vspace{-3mm}
    \caption{Distances between each pair of robots for the narrow corridor lab experiment.}
    \label{fig:robot_robot_dist}
\end{figure}

Fig. \ref{fig:corridor} shows snapshots from an experiment in a constructed environment containing a narrow corridor, intended to evaluate the formation's ability to deform. As the formation enters the corridor in subfigure b), it organically transforms into a line as a consequence of the virtual springs. The line shape is maintained through the bends in the corridor (subfigure c), after which it reverts to the triangular base configuration (subfigure d). Fig. \ref{fig:robot_obstacle_dist} distances between the center of each robot and its closest obstacle for the corridor experiment are shown. It can be seen that all distances maintain above the inflation radius $\tilde{r}_\mathrm{max}=\SI{0.30}{m}$ (approximately half the length of a Go1) at all times with a good margin. Similarly, in Fig. \ref{fig:robot_robot_dist}, which shows the distances between each pair of robots, it can be seen that the inter-robot distances remain well above $2\tilde{r}_\mathrm{max}$, verifying that inter-robot collisions are avoided. 
It is observed that before the formation enters the narrow corridor ($t = 7.3$ s), all inter-robot distances remain close to the relaxed spring length $\tilde{l}_0 = \SI{1.15}{m}$. As the formation deforms organically to navigate through the confined space, these distances temporarily deviate from their nominal values. Once the formation exits the corridor ($t = 35$ s), the inter-robot distances gradually return to the relaxed spring length, indicating the system’s ability to recover its original configuration.

\section{Conclusion}
This paper presents a dynamic online framework for formation path planning and control of multiple quadruped robots based on a role-adaptive leader-follower approach coupled with a system of virtual springs and dampers and a final obstacle avoidance layer. The proposed framework guides the formation along cluttered environments to the final goal while also relaxing the formation in order to pass through safely. The proposed approach is evaluated in a Gazebo simulated environment along with physical experiments in various test cases involving three quadruped robots. This framework represents a significant step toward enabling cooperative, flexible, and obstacle-adaptive multi-quadruped robot systems for real-world deployment.




\bibliography{referencesWithoutUrls}             
                                                   







\end{document}